\documentclass[annual]{acmsiggraph}            

\usepackage[scaled=.92]{helvet}
\usepackage{times}
\usepackage{epsfig}
\usepackage{graphicx}
\usepackage{parskip}
\usepackage{amsmath}
\usepackage{overpic}
\usepackage{color}
\usepackage{soul}
\setstcolor{red}
\usepackage{parskip}
\usepackage{wrapfig}
\usepackage[labelfont=bf,textfont=it]{caption}

\newcommand{\etal}{\textit{et~al}.}
\newcommand{\vnudge}{\vspace{-.1in}}
\newcommand{\mypara}[1]{\paragraph{#1.}}
\long\def\symbolfootnote[#1]#2{\begingroup%
\def\thefootnote{\fnsymbol{footnote}}\footnote[#1]{#2}\endgroup}
\newcommand{\spacerr}{\hspace{.25in}}
\newcommand{\spacers}{\hspace{.4in}}
\newcommand{\figures}[1]{figures/#1}
\newcommand{\rgbd}{RGBD}

\TOGonlineid{0362}
\TOGvolume{}
\TOGnumber{}
\TOGarticleDOI{}
\TOGprojectURL{}
\TOGvideoURL{}

\TOGcodeURL{}

\title{SmartAnnotator: An Interactive Tool for Annotating {\rgbd} Indoor Images}

\author{Yu-Shiang Wong$^1$ \spacerr Hung-Kuo Chu$^1$ \spacerr Niloy J. Mitra$^2$ \vspace{.07in} \\  %
$^1$National Tsing Hua University \spacers $^2$University College London}
\pdfauthor{James Hung-Kuo Chu}


\begin{document}

 \teaser{
   \includegraphics[width=\textwidth]{\figures{teaser}}
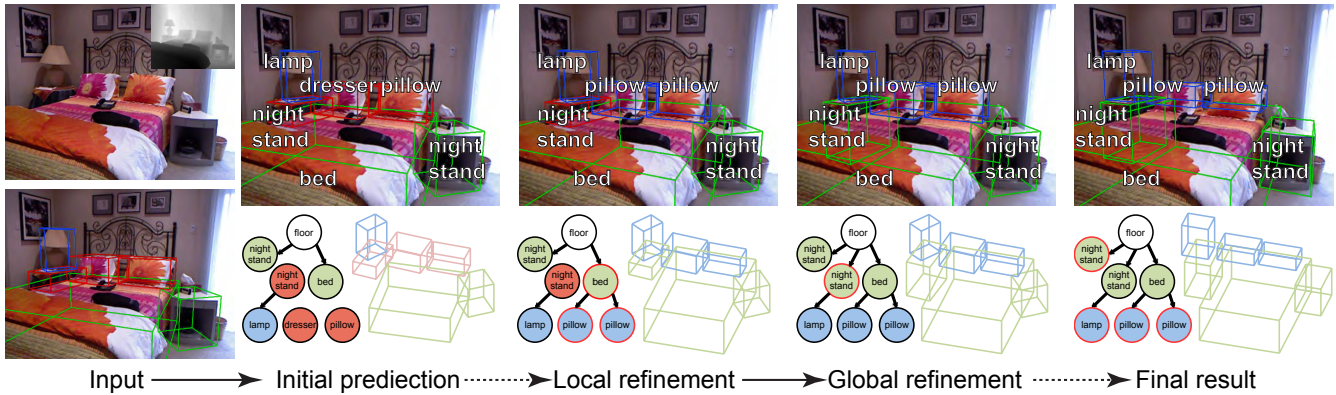
   \caption{We present SmartAnnotator, an interactive tool to facilitate annotating {\rgbd} images. The system starts by predicting labels using learned priors while the user provides supervision by accepting the predicted bed label (the first dashed arrow). The system then locally refines the bed geometry, and globally infers support relationships with
   pillows and resolves occluded parts of nightstand. The user then simply approves all labels to end the annotation process (the second dashed arrow).  SmartAnnotator enables the user to effortlessly annotate this {\rgbd} image simply by two clicks (one to confirm the bed;  another to approve all) and took less than 5 secs in this example.}
   \label{fig:teaser}
   \vnudge
 }

\maketitle


\begin{abstract}

RGBD images with high quality annotations in the form of geometric (i.e., segmentation) and structural (i.e., how do the segments are mutually related in 3D) information 
provide valuable priors to a large number of scene and image manipulation applications. While it is now simple to acquire RGBD images, annotating them, automatically or manually, remains challenging especially 
in cluttered noisy environments. 
We present SmartAnnotator, an interactive system to facilitate annotating RGBD images. 
The system performs the tedious tasks of grouping pixels, creating potential abstracted cuboids, inferring object interactions in 3D, and comes up with various hypotheses. The user 
simply has to flip through a list of suggestions for segment labels, finalize a selection, and the system updates the remaining hypotheses. As objects are finalized, the process speeds up with fewer ambiguities to resolve. 
Further, 
as more scenes are annotated, the system makes better suggestions based on structural and geometric priors learns from the previous annotation sessions.
We test our system on a large number of database scenes and report significant improvements over naive low-level annotation tools.

\end{abstract}




\keywordlist


\TOGlinkslist



\section{Introduction}
\label{sec:intro}
%
Images with high quality semantic annotations provide rich source of training data for a variety of supervised and semi-supervised algorithms, both in computer graphics and computer vision.
For example, in scene understanding, algorithms  extract cues from the annotated datasets to learn dominant relationships between object label and image contents. The trained models are then used as {\em priors} for manipulation, reconstruction, synthesis, etc.
Beyond use for training priors, such annotated datasets also provide qualitative and quantitative groundtruth for evaluating segmentation and labeling algorithms.
%
%
Gathering such data relies on heavy manual effort with the user annotating images one at a time. The process is a tedious and time-consuming task, resulting in errors by the (tired) users.
State-of-the-art web-based image annotation tools (e.g.,~LabelMe~\cite{Russell:08:Label}) simplify the process by facilitating collaborative annotation and offering  easy-to-draw interfaces. The users, however, still have to manually prescribe polygonal segments and type in object labels individually.
The situation is even worse when dealing with challenging indoor images usually containing cluttered objects with complex boundaries and heavy occlusion.
%
%

{\rgbd} sensors (e.g., Microsoft Kinect) provide easy and affordable synchronized color and depth data.
Not surprisingly, in the context of scene understanding, priors learned by utilizing such depth from correctly annotated {\rgbd} data result in dramatic performance gains.
The annotation process, which inherits the problems of the image setting, is further complicated since the raw depth data is often noisy, contains outliers, and suffers from occlusion.
Thus, while properly annotated depth data can be invaluable, the manual annotation process itself is difficult posing a severe bottleneck.

Existing papers either work on 2D segments and treat depth data as an additional feature channel~(e.g.,~\cite{silberman:12:indoor,Ren:12:RGBD}), or reason about scene structure on the 3D patches using point cloud~(e.g.,~\cite{koppula:11:semantic}).
Xiao~\etal~\shortcite{Xiao:13:SUN3D} introduced SUN3D, an annotated database of full 3D places, that integrates depth data across multiple video frames into a full 3D point cloud model.
Yet, it still inherits the data quality issue and how it could be exploited to reason the 3D structure of scene remains to be explored.
Unfortunately, we still lack a smart annotation tool that utilizes depth data and simplifies the users' task of annotating. 
%

We present an interactive tool to annotate {\rgbd} indoor images.
As output the system provides both image and scene level segmentation, segment labels, and structural relationships (e.g., contact, on-top, etc.) among the segments.
%
%
This is achieved via combining a novel scene labeling scheme with object annotating tasks so that they mutually assist each other.
The system, in the background, performs the tasks of computing segmentations, predicting the object labels, and inferring the 3D structure of scene. The user simply supervises the process by optionally providing initial scribbles
and then progressively accepting suggestions from the system. Thus the user only selects among ordered suggestions (e.g., if a shown box is `bed' versus `cabinet') while the system updates its {\em understanding} of the scene and proposes refined suggestions, both in terms of updated tags and segments, for the remaining objects. At any point the user can `approve all' to finish the process.  Figure~\ref{fig:teaser} shows stages from such a session (see also supplementary video).

%
%
Reconstructing a detailed 3D model from a single image remains an ill-posed problem even in presence of depth cues.
In the context of indoor scenes comprising of man-made objects, we parse the scenes into simple room layouts and a collection of approximate cuboids using both color and depth information.
Therefore, both geometric and structural priors (e.g., size, spatial and support relationships among objects) are exploited via reasoning on this concise 3D representation.
Note that the prior progressively gets richer. Specifically, earlier scenes get encoded as prior, which in turn simplifies annotation of subsequence {\rgbd} images. The task becomes simpler as the user processes more scenes.

Our system works in two phases:
First, in a {\em learning phase}, we bootstrap the scene labeling using a handful of labeled {\rgbd} images (10 scenes) with properly refined 3D structures from where the algorithm learns geometric and structural priors.
Second, in the key {\em annotating phase}, the input {\rgbd} image is parsed into a 3D structure followed by reasoning possible support relationships and predicting the labels using the estimated cuboids and learned models, respectively.
The user browses and confirms suggestions, i.e., labels, proposed by the system, while the algorithm immediately refines dimension of cuboids, segmentations, support relationships, and re-estimates the labels in response to user interactions.
The process continues until all the objects are properly labeled. The system augments the existing dataset by appending the newly annotated image.
We evaluate the effectiveness of our system on benchmark {\rgbd} dataset (126 scenes)
both in  terms of performance and quality in annotation.
With our tool, we demonstrate that annotating a {\rgbd} image could be done in a few user clicks and typing without losing the accuracy in the annotated data (see supplementary material and video).

\vnudge
\mypara{Contributions} In summary, our main contributions include:
\begin{itemize}
\item an interactive annotation tool that enables user to annotate {\rgbd} indoor images quickly and accurately; \vspace{-.05in}
\item combining the object annotating tasks with a novel scene labeling that exploits geometric and structural priors via reasoning on the 3D volumetric representation of {\rgbd} images using a room layout and cuboid relationships; and \vspace{-.05in}
\item a context-driven 3D scene structure refinement to automatically adjust the dimension and support relationships of cuboids according to user annotation.
\end{itemize}

\section{Related Work}
\label{sec:relatedWork}
\mypara{Traditional image annotation}
The ability to collect a large amount of annotated images is crucial for applications in computer vision.
Russell~\etal~\shortcite{Russell:08:Label} developed a web-based image annotation tool, called LabelMe, to collect a large dataset of labeled images.
It provides user an easy-to-use drawing interface to annotate object at different level of complexity and allows a large population of users to work collaboratively.
Nowadays, LabelMe or variants are popular forms of image annotation tool to serve many state-of-the-art image datasets~\cite{Russel:09:lT,Xiao:10:SUNDB}.
However, such 2D annotation tool at most provides object labels and their image segments, even the image itself contains rich depth data from {\rgbd} sensors~\cite{silberman:12:indoor,Xiao:13:SUN3D}.
In this paper, we utilize the depth data and propose an interactive tool to facilitate annotating {\rgbd} indoor images with rich data in the context of both 2D image contents and 3D structure of the scene.
To the best of our knowledge, ours is among the first works which aims at annotating {\rgbd} images and we believe a dataset with properly annotated depth data can be invaluable to advanced scene understanding.

\mypara{Indoor scene labeling}
Indoor scene labeling has been extensively studied in the field of scene understanding.
While a huge body of work has focused on designing good local image features (e.g., SIFT and HOG), the growing popularity of depth sensors has further renewed the perspective of traditional approaches to incorporate 3D features.
Silberman and Fergus~\shortcite{silberman:11:indoor} treated depth as an additional channel and extracted image features from both color and depth images to perform the labeling task.
Later, Ren~\etal~\shortcite{Ren:12:RGBD} further investigated combining rich {\rgbd} features and applied context modeling using MRFs and a sgementation tree to obtain dramatic performance gains.
Instead of treating depth data as an additional feature channel, Koppula~\etal~\shortcite{koppula:11:semantic} extracted various geometric and contextual features from 3D patches using point cloud.
In contrast, we propose a novel scene labeling algorithm that reasons on the 3D structure of the scene, which consists of a simple room layout and a collection of cuboids inferred from depth data.
%
This goes beyond the scope of traditional scene labeling.

\mypara{Image-based 3D scene modeling}
Reconstructing detailed 3D scenes from images has been widely investigated both in computer graphics and computer vision.
Although it is well-known that a properly constructed 3D scene can be useful in versatile applications including scene understanding~\cite{Gupta:10:BWR,Hedau:10:box} and image manipulation~\cite{Karsch:11:RSO,zheng:12:iImages}, missing depth information in camera projection makes the problem ill-posed.
Hence, methods usually rely on detecting image features or high-level annotation from user to guide the reconstruction and simplified the 3D representation in a form of popup planar segments~\cite{Russel:09:lT,Saxena:09:MLS} or approximate primitives such as cuboid~\cite{Hedau:10:box,Gupta:10:BWR}.
For example, according to Manhattan world assumption~\cite{Coughlan:03:Manhattan}, Hedau~\etal~\shortcite{Hedau:10:box} extracted vanishing points from straight line cues and parsed the geometry of a room using 3D oriented boxes.
While Gupta~\etal~\shortcite{Gupta:10:BWR} used user annotation and applied geometric and physical constraints on a 3D parse graph, and modeled the scenes using axis-aligned blocks.

Motivated by the availability of depth data from {\rgbd} sensors, a recent progress has been made in modeling scenes from {\rgbd} images.
However, the high frame-rate in acquisition comes at the cost of data quality and parts of data are easily lost due to occlusion.
%
%
Jiang and Xiao~\shortcite{Jiang:13:cuboid} formulated the problem of matching cuboids to segments as minimizing the local fitting error (e.g., minimize distance from 3D points to visible faces of cuboid) via evaluating global structure constraints (e.g., small occlusion among nearby cuboids).
Further, Jia \etal~\shortcite{jia:13:3d}  incorporated support and stability inference into the matching pipeline to obtain plausible cuboid configuration.
Our approach of fitting cuboids is mainly inspired by previous ones and is adapted in a simplified formation to facilitate interactive performance.
%


\begin{figure*}[!t]
    \centering
    \includegraphics[width=\textwidth]{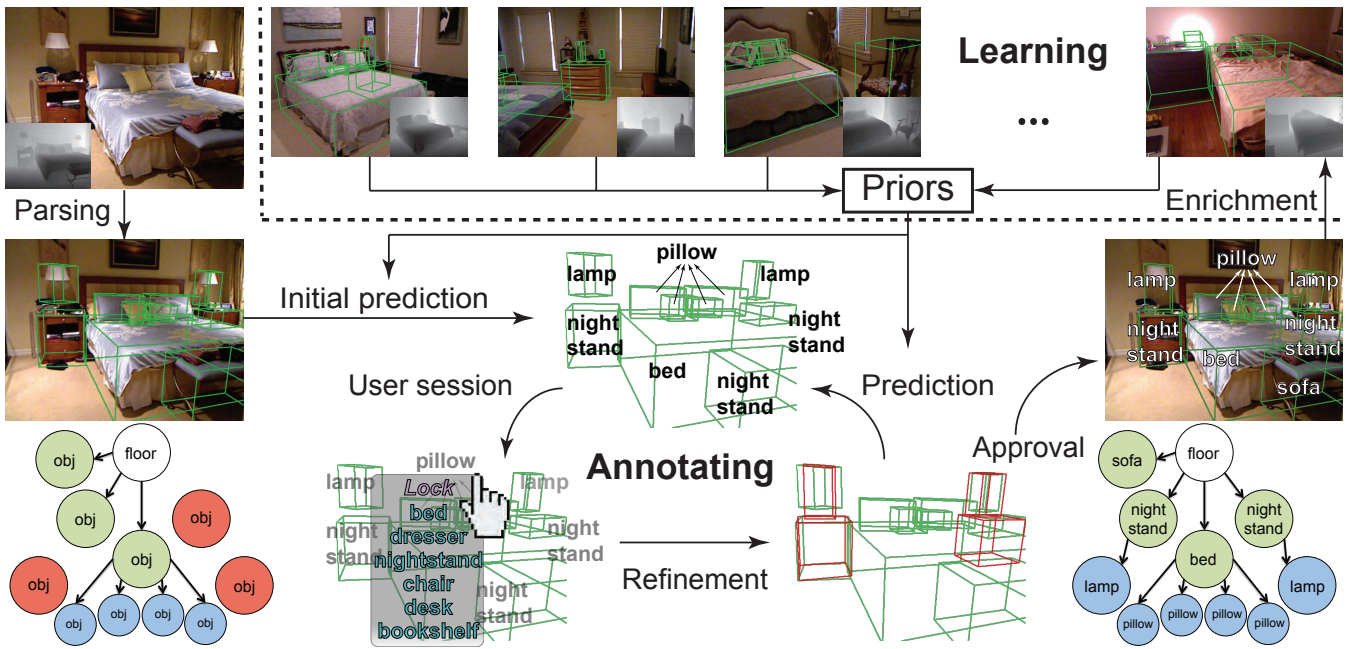}
    \caption{System overview: Input to the learning phase is a small amount of labeled {\rgbd} images with properly refined 3D structures (highlighted cuboids) from which the priors are learned. In the annotating phase, system takes an input {\rgbd} image, constructs its 3D structure which is encoded as a structure graph, and predicts object labels using the learned priors. User supervises the process by progressively accepting suggestions from the system while system then automatically refines the structure graph and re-predicts object labels. The process iterates until user approving all labels and append new image to database to enrich the priors.}
    \label{fig:pipeline}
    \vnudge
\end{figure*}

\section{Overview}
\label{sec:overiew}
To annotate a {\rgbd} indoor image with 2D/3D information including image and scene level segmentation, segment labels, and structural relationships, our system works in two phases: learn and reason on {\rgbd} data (learning phase) followed by utilizing the learned priors to assist user in annotation (annotating phase), as illustrated in Figure~\ref{fig:pipeline}.

In the {\em learning phase}, the system takes a handful of labeled {\rgbd} indoor images (10 scenes) as input.
Each {\rgbd} indoor image is parsed into a 3D structure of scene comprising of a room layout (e.g., floor and walls), and a collection of 3D cuboids to represent objects (Section~\ref{sec:modeling}).
In order to generate a baseline training data, we expect minimal user intervention to assist the construction of 3D structures in images where the quality of data is poor or missing due to occlusion.
Such properly refined 3D structure is then represented as a {\em structure graph} which is the core processing unit in our system as well as the target data of annotation (Section~\ref{sec:SG}).
The learning algorithm bootstraps by reasoning on structure graphs and learning geometric and structural priors (Section~\ref{sec:learning}).

In the object {\em annotating phase}, system takes an input of {\rgbd} image, parses the image into a 3D structure of the scene and constructs a structure graph accordingly.
We formulate the problem of label prediction as evaluating a joint probability function, which is modeled based on the structure graph.
%
We employ a greedy approach to incrementally infer a list of suggestions for each object via traversing the structure graph and evaluating the probabilistic function using the learned priors. (Section~\ref{sec:predict}).
The control is then taken by user who is responsible for supervising the system.
User interacts with system through an interface that allows he/she to guide the system just by confirming, reordering or overriding (e.g., typing) the suggestions proposed by system (Section~\ref{sec:user}).
While system, in background, automatically updates its understanding to the scene in accordance with user's actions and refines the 3D structure of scene (e.g., resolving ambiguity and occlusion), object segments, and re-predicting labels for the remaining objects (Section~\ref{sec:refine}).
The process iterates until user approves all the predicted labels.
Then, we progressively get richer priors by augmenting the existing labeled dataset with the newly annotated images and retraining the priors, as demonstrated in Section~\ref{sec:eval}.
%

\begin{figure*}[!t]
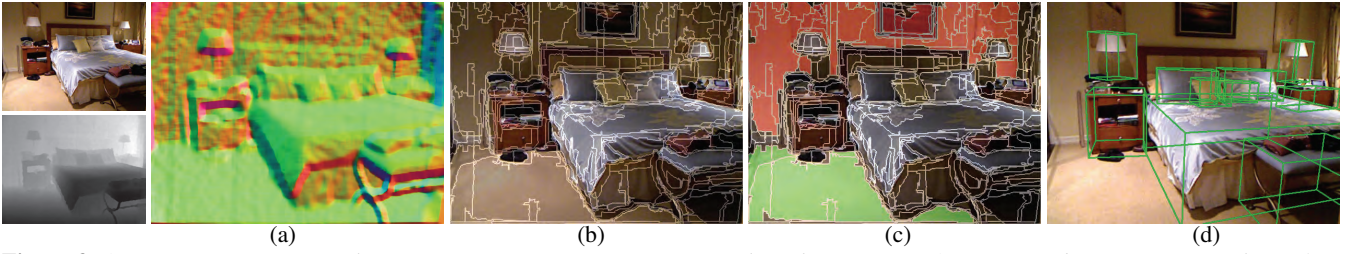

    \centering
    \begin{overpic}[width=\textwidth]{\figures{3D_structure_pipeline}}
    \put(20,-1.3){(a)}
    \put(43,-1.3){(b)}
    \put(66,-1.3){(c)}
    \put(89,-1.3){(d)}
    \end{overpic}
    \caption{3D structure parsing pipeline. Given an input {\rgbd} image, our algorithm parses its 3D structure by (a) computing the surface normals, (b) computing over-segmentation using colour and normal images, (c) extracting the room layout (floor and wall are highlighted in green and red) and (d) fitting cuboids.}
    \label{fig:structure}
    \vnudge
\end{figure*}

\section{Modeling Scene Structure}
\label{sec:modeling}
The core of system builds on learning and annotating the geometry and structural relationships by reasoning a 3D interpretation of the underlying indoor scene.
Reconstructing a detailed 3D model from a single image, however, is still an ill-posed problem even in presence of depth cues.
Despite the difficulty, the recent research efforts in scene understanding~\cite{Gupta:10:BWR,Hedau:10:box} and image manipulation~\cite{Karsch:11:RSO,zheng:12:iImages} have shown that rich geometric, structural and contextual information are encoded in a simplified 3D representation using approximate cuboids.
We design a lightweight algorithm to parse a {\rgbd} image into a 3D representation comprised of a simple room layout with floor and walls, and a collection of cuboids for objects.
We adopt an over-segmentation of the {\rgbd} image followed by fitting planes and cuboids to image segments to infer the 3D structure of scene.
Figure~\ref{fig:structure} shows the pipeline of algorithm.

First, we compute the surface normals at each pixel by fitting a least-square plane using the neighboring pixels, and render the result to a color-coded normal map (see Figure~\ref{fig:structure}(a)).
Then a graph-based image segmentation~\cite{Felzenszwalb:04:EGI} is applied to over-segment both the RGB image and normal map.
The initial segmentation is generated by superimposing two over-segmented images and taking the union of their segment boundaries (see Figure~\ref{fig:structure}(b)).
For each segment, a 3D plane is estimated by applying a RANSAC method on the corresponding 3D points.
We next describe a semi-automatic algorithm to model the room layout and fit cuboids based on these image segments and corresponding 3D points and planes.

\mypara{Notations}
We denote the initial segmentation as $S=\{s_1,...,s_{N_s}\}$, where each segment $s_i=\{\textbf{I}_i, \textbf{X}_i, \textbf{p}_i, \textbf{n}_i\}$ encodes the image pixels, 3D points, 3D plane and normal to the 3D plane, respectively.
Segments $s_i$ and $s_j$ are parallel (orthogonal) if the angle between $\textbf{n}_i$ and $\textbf{n}_j$ is within (above) a tolerance angle of $a_T$ ($90 - a_T$).
The distance between $s_i$ and $s_j$ is defined as Min$(D(\textbf{X}_i, \textbf{p}_j), D(\textbf{X}_j, \textbf{p}_i))$, with $D(\textbf{X}, \textbf{p})$ calculates the closest distance between 3D points $\textbf{X}$ and plane $\textbf{p}$.
We define $s_i$ and $s_j$ are {\em coplanar} if $s_i$ and $s_j$ are parallel and distance between them is within a threshold of $d_T$.
Unless otherwise mentioned, we use $a_T=30$ degrees and $d_T=15$ cm across the paper.

\mypara{Extracting Room Layout}
We extract the floor by using the gravity information from the accelerometer data of {\rgbd} sensor to infer the floor segments.
First, we search for a seed segment which is orthogonal to gravity direction and is at the lowest boundary of the scene in that direction.
Then we iteratively merge this seed segment with other coplanar segments and fit a new 3D plane to the joint 3D points using RANSAC method, until no more candidate segments could be found.
All the segments in the final collection are labeled "floor" and is denoted as $S_f$. 

Similar to finding the floor, we draw the problem of extracting the walls on determining a seed segment for each wall and progressively growing from this seed segment.
Assume a room has two dominant and orthogonal walls, we adapt the approach from Silberman~\etal~\shortcite{silberman:12:indoor} by collecting segments which are orthogonal to floor and forming a set of candidate wall pair $(\textbf{n}_i, \textbf{n}_j)$, where $\textbf{n}_i$ and $\textbf{n}_j$ are normals of two orthogonal segments.
We evaluate the score of each candidate pair as follows:
\begin{equation}
E_w(\textbf{n}_1, \textbf{n}_2) := \sum_{i=1}^{2} \sum_{j=1}^{N_p} \; \exp(-(\textbf{n}^p_j \cdot \textbf{n}_i)^2)
\end{equation}
where, $(\textbf{n}_1, \textbf{n}_2)$ is a candidate pair, $N_p$ denotes the number of 3D points from the collected segments and $\textbf{n}^p_i$ is the surface normal of a 3D point.
We choose the candidate pair that has the largest score, use the normals to find seed segments for walls and grow from the seeds to form two labeled wall segments $S_{w_0}$ and $S_{w_1}$.
Figure~\ref{fig:structure}(c) shows an example of extracted room layout.

\mypara{Fitting Cuboids}
%
%
%
For the rest of unlabeled segments, our goal is to fit a cuboid to each object segment.
In order to specify individual object segment, our system provides a drawing interface for user to roughly scribble on the image using two kinds of strokes, one indicates foreground while the other is background.
Then we run the GrabCut~\cite{Rother:2004:GIF} to generate a foreground mask and force the mask boundary to be consist with the initial segmentation to finish an object segment.
%
%

The orientation of a 3D bounding box is determined by two perpendicular normals.
To simplify the problem, we assume that all the cuboids are standing up-right against the floor or stacked on top of other cuboids.
Thus,  fitting  only needs to determine one dominant normal while fixing the other to the normal of floor.
Given a set of 3D points from the object segment, we project them to the floor as 2D points, excluding 3D points with surface normal perpendicular to floor.
We gather 2D points that are close to the boundaries of convex hull of the projection and fit a line to these 2D boundary points using RANSAC.
The vector that is perpendicular to this line gives the second dominant normal of box.
And the dimension (or size) of the box is determined by calculating a minimum volume of box that extends to the boundaries of input 3D points given the box orientation.
Figure~\ref{fig:structure}(d) shows an example of the process.

\mypara{User Intervention}
If the automatic detection fails (e.g., missing floor or walls, more than two walls, and incorrect object segments), we expect the user to intervene.
For example, in the construction of room layout, user could initially specify the seed segments for floor and walls by clicking on the image segments or scribble on the image to specify and refine the floor, wall and object segments.

\begin{figure*}[t]
    \centering
    \includegraphics[width=\textwidth]{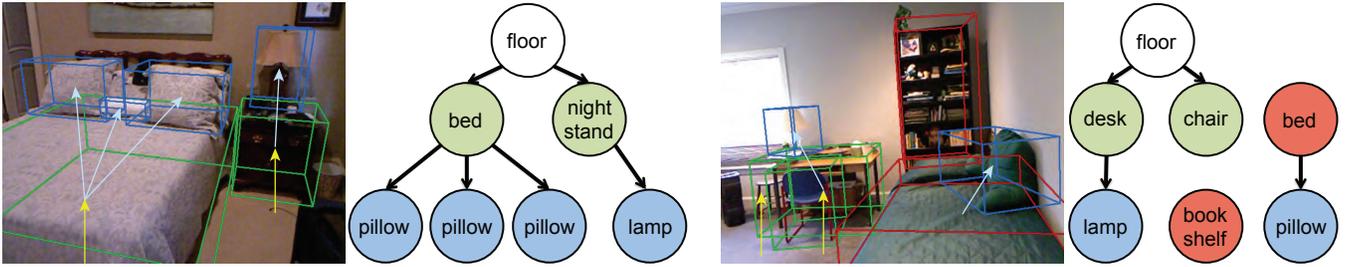}
    \caption{Examples of structure graphs illustrated in 3D and 2D. The ground objects are represented by green cuboids (nodes) while the floating objects are in red. The arrows indicate support relationships between objects (in light blue) and relative to floor (in yellow).}
    \label{fig:SG}
    \vnudge
\end{figure*}


\subsection{Structure Graph}
\label{sec:SG}
We now introduce the {\em structure graph} to encodes geometric and structural information from the 3D structure of scene as illustrated in Figure~\ref{fig:SG}.
A structure graph (or SG hereafter) is represented as a {\em directed} graph $G:=(V,E)$, with a root node $v_f \in V$ denoting the floor and each node $v_i \in V$ indicating an object where $v_i=\{S_i, c_i, r_i, w^c_i, w^a_i\}$ encodes the information of object segment ($S_i$), estimated cuboid ($c_i$), the projection of $c_i$ on the floor ($r_i$), and spatial relationships to the room layout ($w^c_i$ and $w^a_i$).
Each directed edge $e_{ij} \in E$ describing a relationship of $v_j$ is supported by $v_i$.
Therefore, given a parsed 3D structure of scene, we construct the corresponding SG by using its object segments and cuboids, and reason on the structural relationships as listed below:

{\em (i) Spatial relationships: }
The spatial relationships of object $v_i$ with respect to room layout which are two binary values $w^c_i$ and $w^a_i$ indicating whether or not the cuboid $c_i$ contacts and aligns to the nearest wall, respectively.
To model such relationships, we determine the "back" face of every cuboid which is the face closest to any wall in the room, excluding top and bottom faces.
A cuboid $c_i$ is defined as contacting the wall ($w^c_i=1$) if the distance between its back face and the nearest wall is within a threshold ($d_T$).
And a cuboid $c_i$ is defined as aligning to the wall ($w^a_i=1$) if its back face is parallel to the nearest wall within a tolerance angle ($a_T$).

{\em (ii) Support relationships:}
Based on the assumption that all the cuboids are standing up-right against the floor or stacking on top of other cuboids, we infer the support hierarchy among objects as follows.
For each object $v_i$, we add a directed edge $e_{fi}$ to $E$, indicating $v_i$ is supported by floor, if the bottom face of $c_i$ is close to floor ($\leq d_T$).
We define the relationships of $v_i$ is supporting $v_j$ as follows:
(i) The distance between the top face of $c_i$ and the bottom face of $c_j$ is within a threshold ($d_T$), and given two projected bounding rectangles $r_i$ and $r_j$, the center point of $r_j$ falls inside $r_i$ or the intersecting area of $r_i$ and $r_j$ is greater than $30\%$ area of $r_j$.
%
(ii) To account for the situation where an object $v_j$ is supported by another non-convex object $v_i$, we define another support criteria as $c_j$ is completely contained in $c_i$.
%
We add a directed edge $e_{ij}$ to $E$ if one of the above criteria is true.
For simplicity, we assume each object is supported at most by one object (or floor), and if there are more than one supporting objects, we choose the most probable one with largest intersecting area.
Lastly, we define two sets, $V_g$ and $V_f$, to represent objects that are supported by floor and have no supporting parent, respectively.

\section{Learning Phase}
\label{sec:learning}
Learning informative feature priors plays the key role to a robust scene labeling.
We found traditional appearance-based features (e.g., SIFT or HOG) to be insufficient to effectively distinguish image patches apart due to the factor that objects in indoor scenes come in a variety of categories with large variation in appearance
%
In contrast, despite the appearance, human can immediately recognize each object in a room merely by its geometry (e.g., bed has larger base than bookshelf but is shorter in height), its spatial distribution in the room (e.g., bed and bookshelf are placed on the ground and against to wall) and its support relationships (e.g., pillow is frequently on top of bed while lamp is on top of desk).
We propose a scene labeling algorithm to learn geometric and structural priors via reasoning on 3D structures inferred from underlying scene, and exploit the learned models to assist user in smart annotation. 
%

\subsection{Learning Priors}
\begin{figure}[b!]
\vnudge
    \centering
    \includegraphics[width=\columnwidth]{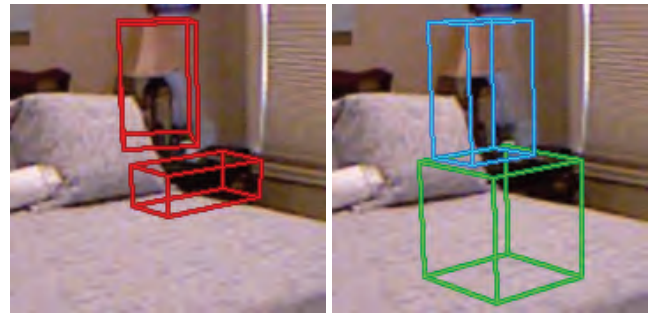}
    \caption{(Left) An incorrect 3D structure. The size of both lamp and nightstand are bad due to occlusion. (Right) We apply local refinement (see Section~\ref{sec:refine}) to fix the size and orientation of cuboids.}
    \label{fig:manual}
\end{figure}

\mypara{Dataset}
We bootstrap the learning process with 10 labeled {\rgbd} indoor images which  contained detailed object labels and segments manually annotated by user (e.g., via LabelMe).
Based on the given annotations, each image is parsed into a 3D structure using the algorithm described in Section~\ref{sec:modeling}.
However, the raw depth data acquired from a single view often suffers from artifacts like noisy data with outliers and occlusion among objects, hence resulting incorrect 3D structures (see Figure~\ref{fig:manual}(left)).
We manually refine, if necessary, the dimension and orientation of each cuboid to improve overall spatial relationships and resolve the structural ambiguity (see Figure~\ref{fig:manual}(right)).
Besides providing baseline training data for learning reliable priors, such local refinement also plays a key role in our system to automatically and progressively update the target SG during the run time annotation (see Section~\ref{sec:refine}).

Given properly refined 3D structures, we construct the corresponding SGs which are then served as initial training data.
In the following, we elaborate how to learn the geometric and structural priors from SGs regarding to a list of indoor object categories, denoted as $O=\{o_1,...o_k\}$.
%


\mypara{Geometric model}
The design of indoor objects is close related to their functionality for supporting human activities.
Therefore, the 3D size of object can be an effective cue to distinguish different object categories.
For example, the bed for sleeping is large in base and low in height comparing to bookshelf for storing which is smaller in base and longer in height.
Thus, we model distribution of 3D size of object category using 2D distribution of the bottom face area and height of cuboids from SGs, and then learn the geometric prior using a multi-class probability SVM~\cite{Chang:11:libsvm}.
The final geometric model is denoted as $\mathcal{P}_g(o,v)$ which return a value in the range [0, 1.0], indicating how likely an object $v$ is classified to category $o$ given the 3D size of object's cuboid.
%

\mypara{Data enrichment}
The set SGs being small, we might suffer from the insufficient (or zero) samples for each object category, resulting in an ineffective (or invalid) probabilistic model.
Hence, we introduce a separate text-based dataset from IKEA\textsuperscript{\textregistered} to enrich the samples of objects.
Specifically, for each object category, in addition to samples from SGs, we add extra 20 samples by fetching the size specification from the IKEA\textsuperscript{\textregistered} website and then apply random jitter to obtain a total amount of 50 samples.
We jitter object area by $\mathcal{N}(0,100cm^2)$ and height by $\mathcal{N}(0,10cm)$ with $\mathcal{N}(.)$ indicates a normal distribution.

\mypara{Spatial model}
In interior design, the arrangement of most objects (e.g., furniture) is mainly dominated by the geometry of room layout.
To model the spatial relationships of object with respect to the room layout, the commonly used metrics are measuring objects' relative distance and orientation to the walls of the room~\cite{Merrell:11:IFL,Yu:11:furniture}.
While the existing approaches aim to learn such priors to guide the object rearrangements, we encode the relationships to the walls as spatial constraints which are used to guide the local refinement in Section~\ref{sec:refine}.
We define two spatial constraints for each object category as follows.
%
%
An object category $o_i$ is tagged as contacting (aligning to) wall if the ratio of object $v_j=o_i$ and $w^c_j=1$ ($w^a_j=1$) to all the occurrence of $v_j=o_i$ in SGs exceeds a threshold (0.7).
We further classify each object category into two sets, $O_c$ and $O_p$, according to the tagged spatial constraints of contacting wall and aligning to wall, respectively.

\mypara{Support model}
In the context of indoor scene modeling, the support relationships are proven to be a strong cue describing a local structure between two objects~\cite{Yu:11:furniture,jia:13:3d,Fisher:12:ESO}.
For example, both pillow and lamp tend to be supported by bed and desk while bed and desk are supported only by the floor.
In this paper, we model the support relationship of two object categories, $o_i$ and $o_j$, by simply counting the frequency of $e_{ab} \in$ SGs, where $v_a=o_i$ and $v_b=o_j$, among all the co-occurrence ($v_a$, $v_b$) in SGs.
We denote the frequency as $\mathcal{P}_s(o_i,o_j)$ which indicates the likelihood of $o_i$ is supporting $o_j$.
As for the support relationship between floor and object, we model it as a support constraint.
Specifically, we create a set $O_s$ which includes object category $o_i$ that satisfies the criteria, $\mathcal{P}_s(floor, o_i) \geq 0.7$.
%

\begin{figure*}[t]
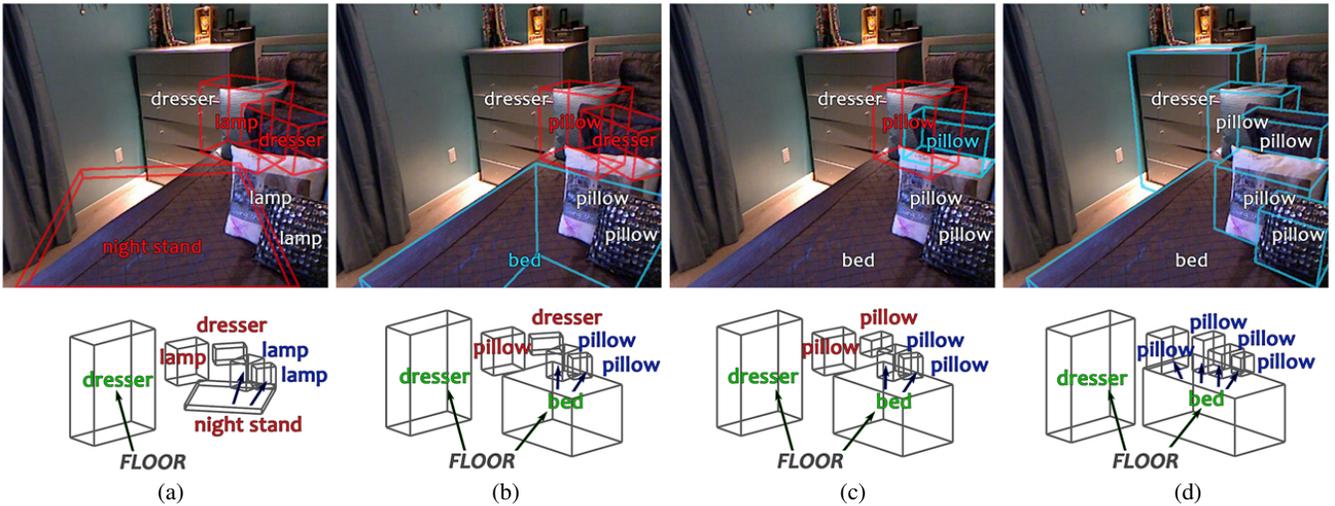

    \centering
    \begin{overpic}[width=\textwidth]{\figures{SG_refine}}
    \put(12,-1.0){(a)}
    \put(37,-1.0){(b)}
    \put(63,-1.0){(c)}
    \put(88,-1.0){(d)}
    \end{overpic}
    \caption{Progressive structure refinements. (a) The initial structure and predicted labels. (b) User re-orders from nightstand to bed and system performs local refinements to improve the dimension and orientation of bed's cuboid. (c) User further assists in resolving the ambiguity arise from nearby object and approves all the labels. (d) While system globally infers the relationships among objects, and thus expanding bed to support two floating pillows.}
    \label{fig:SG_refine}
    \vnudge
\end{figure*}

\section{Annotating Phase}
\label{sec:annotate}
Our goal in this section is to annotate a given {\rgbd} image with both image and scene level segmentation; segment labels; and geometric and structural relationships among segments.
The learned models of geometric and structural priors ($\mathcal{P}_g$ and $\mathcal{P}_s$), and spatial and support constraints ($O_c$, $O_p$ and $O_s$) are used to facilitate smart annotation, as described next.

The input to the annotating system is a {\rgbd} image that is segmented, parsed and encoded as a SG.
Our system starts by predicting the object labels using the learned priors on a joint probability function inferred from the input SG.
Unfortunately, as mentioned earlier, the data quality issues often lead to incorrect structure of SG, thus resulting in inaccurate labeling results.
We correct this by inferring a list of ordered suggestions for each object, instead of predicting only one, by greedily evaluating the joint probability function based on the structure of SG.
User then takes control and provides supervision simply by confirming, reordering or overriding (e.g., typing) the suggestions proposed by system.
The system then updates its understanding to the scene in terms of refining the structure of SG, object segments, and re-predicting labels for the remaining objects.
User can `approve all' to finish the process at any point of time.

\subsection{Label Prediction}
\label{sec:predict}
Given a SG $G$, we formulate the problem of predicting labels for objects $V = \{v_1,...v_{n}\}$ as a maximum a posteriori (MAP) inference problem that aims at finding the most probable assignment of object categories $L^\ast = \{l^\ast_1,...,l^\ast_{n}\}, l^\ast_i \in O$ and is defined as:
\begin{equation}
\{L^\ast\}=\arg \max_{L} P(L|G)
\end{equation}
where, $P(L|G)$ is a joint probability function defined on SG,
\begin{equation}
\label{equ:prob}
P(L|G)=\prod_{\forall v_i \in V_g,V_f}P(v_i) \cdot \prod_{\forall e_{ij} \in E}P(v_j|v_i).
\end{equation}
By taking $E_p(L|G)=-\log P(L|G)$ and factorizing Equation~\ref{equ:prob} using the prior models ($\mathcal{P}_g$ and $\mathcal{P}_s$), finding the MAP is equal to minimize the energy function,
\begin{eqnarray}
\label{equ:energy}
E_p(L|G)=-\sum_{\forall v_i \in V_g,V_f}\log(\mathcal{P}_g(l_i,v_i))- \nonumber \\
\sum_{\forall e_{ij} \in E}( \log(\mathcal{P}_g(l_j,v_j))+\log(\mathcal{P}_s(l_i,l_j))).
\end{eqnarray}
While the optimal assignment to Equation~\ref{equ:energy} could be found using a variety of well-known numeric methods such as Belief Propagation, we target a different scenario of inferring a list of suggestions for each object.
The advantage of proposing multiple probable labels for each object is two-fold:
(i)~it compensates the issue of prediction accuracy caused by insufficient training data and incorrect structure of SG to some extent; and
(ii)~a properly ordered list would significantly reduce the manual efforts in labeling such that user only selects among the ordered suggestions.
Assume the target number of suggestions is $m$, while a na\"{\i}ve approach to infer such list based on enumerating all possible combinations of labels is prohibitively expensive, we develop an approach to greedily evaluate Equation~\ref{equ:energy} based on the support hierarchy of $G$ in a top-down fashion, starting from ground objects ($v_i \in V_g$) and ending at floating objects ($v_i \in V_f$).

\mypara{Inferring ground objects}
The modeled support constraint $O_s$ presents an effective prior in inferring the suggestions for ground objects.
For each object $v_i \in V_g$, we evaluate the cost function $f_g$ with respect to labels $l_j \in O_s$, where
\begin{equation}
f_g(v_i, l_j)=\log(\mathcal{P}_g(l_j,v_i)) + \sum_{\forall l_k \in O-O_s}\log(\mathcal{P}_s(l_j,l_k)) \cdot \delta(v_i)
\end{equation}
%
%
with $\delta(v_i) = 1$ if $v_i$ is a supporting object and 0 otherwise.
We sort the labels in decreasing cost and suggest the top m labels.

\mypara{Inferring supported objects}
To infer the suggestions for supported object in hierarchy, we exploit the prior knowledge from suggestions of its supporter in hierarchy.
For each object $v_i \in V-(V_g \cup V_f)$ and each label $l_p$ among the suggestions of its parent object in the support hierarchy, we evaluate the cost function $f_s$ with respect to labels $l_j \in O-O_s$ and $\mathcal{P}_s(l_p,l_j) \neq 0$, where
\begin{equation}
f_s(v_i, l_j | l_p):=\log(\mathcal{P}_g(l_j,v_i)) + \log(\mathcal{P}_s(l_p,l_j)).
\end{equation}
Then, we add the label with highest cost to the suggestions.
In the case of none of $l_j$ exists, we simply evaluate the geometric prior $\log(\mathcal{P}_g(l_j,v_i))$ with respect to labels $l_j \in O-O_s$ and select the one with highest cost.
The suggestions are built until all the labels in parent suggestions are visited in order.
We traverse the support hierarchy in a level-order to ensure the suggestions of object is built before visiting its child.
%

\mypara{Inferring floating objects}
Attributing to the missing depth data or occlusion by other frontal objects, the cuboid of floating object is neither supported by the floor nor by any other cuboids.
In both cases, the size of cuboid is wrong and thus using the geometric prior $\mathcal{P}_g$ to infer the suggestions is infeasible.
We use two rules  to predict how likely a floating object would be a ground or supported object by reasoning on the relationships with respect to non-floating objects and generate the suggestions accordingly.

For each object $v_i \in V_f$ and its cuboid $c_i$, we build the suggestions by evaluating the following rules in turn:
(i) If $c_i$ is unlikely to be supported by other nearby non-floating cuboids (see Section~\ref{sec:refine}), we virtually extrude the bottom face of $c_i$ to the floor and infer the suggestions as it is a ground object.
%
%
(ii) Otherwise, $c_i$ could be either supported or occluded by nearby cuboids.
In the former one, we search for the most probable supporting parent $v_p$, virtually extrude the bottom face of $c_i$ to top face of $c_p$, and infer the suggestions $S_1$ based on this predicted support relationship.
In the later, we consider $v_i$ an occluded ground object and infer suggestions $S_2$.
Then we take top $\lfloor \frac{m}{2} \rfloor$ and $m - \lfloor \frac{m}{2} \rfloor$ labels from $S_1$ and $S_2$, respectively, and merge them into the final suggestions.

\subsection{User Session}
\label{sec:user}
After predicting object labels, the system thread enters the user session and waits for feedbacks from user.
To facilitate the annotation process, the interface allows user to interact with system in a quick and intuitive manner.
In display, our system draws cuboid and shows the first label in the ordered suggestions on top of each object.
In annotating, user first clicks on an object and proceeds with one of the following actions:
\begin{itemize}
  \item \textbf{Confirm.}  User clicks the ``Lock" label in the popup menu to confirm the suggestion.

  \item \textbf{Re-order.} User rectifies the predicted label by selecting an alternative label among the suggestions in the popup menu.

  \item \textbf{Type.} None of the suggestions is correct, indicating that the prediction has failed, and user could override the label by typing a new one through a dialog.

  \item \textbf{Approve all.} User clicks the ``Approve all" button to confirm all the labels and finish the annotation process.
\end{itemize}
Each time user performs an action, the system will automatically update its understanding to the scene to respect user's confirmation. 
%

\begin{figure*}[t]
    \centering
    \includegraphics[width=\textwidth]{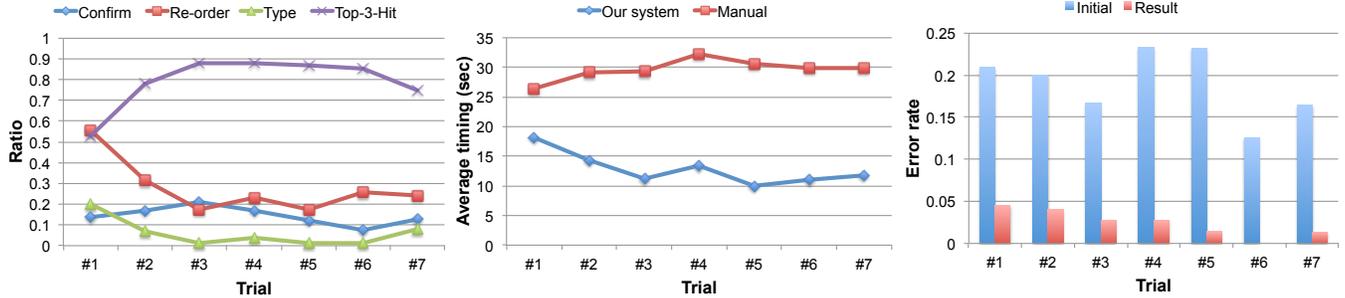}
    \caption{Performance and quality evaluation. Each chart represents (a) the ratio of used manual actions and the accuracy of prediction and suggestions (`Top-3-Hit') during the annotation process, (b) the average annotating time with and without our system, and (c) the quality of support relationships compared to groundtruth in terms of the edge error rate (see Section~\ref{sec:eval}).}
    \label{fig:quanty_eval}\vnudge
\end{figure*}

\subsection{Structure Graph Refinement}
\label{sec:refine}
The bottleneck in obtaining an accurate labeling is in the quality of the input SG $G$, which in itself represents the target data for annotation.
The key challenge is to improve the quality of $G$ and thus boosting the labeling performance for robust annotation.
However, been limited by the quality of data, automatic algorithm at best produces mediocre results and leaving many ambiguities that only human can resolve.
In our cases, the ambiguity comes from floating objects in $V_f$ which might be a supported object or a ground object occluded by some frontal cuboids.
We make use of user annotations and the learned models to progressively refine the structure of $G$ and resolve ambiguity.
Specifically, given the user confirmation, our system performs a joint refinement based on the learned models to locally adjust the dimension and orientation of cuboids, and then globally resolve the ambiguity by examining the inter-object relationships between floating and non-floating objects.

\mypara{Definition}
First, given a cuboid $c_i$ and its expansion $c'_i$, we say $c'_i$ is {\em over-expanded} if $|vol(c'_i)-vol(c_i)| \leq 0.3 \times vol(c_i)$ where, $vol(c)$ is the volume of cuboid $c$, or if there exists any cuboid in the scene that intersects $c'_i$ but not $c_i$.
Second, an object $v_i$ is {\em likely to support} $v_j$ if the following criteria are satisfied:
(i) The distance between the top face of $c_i$ and the bottom face of $c_j$ is within a threshold ($d_T$).
(ii) The closest distance between two projected bounding rectangles $r_i$ and $r_j$ is smaller than a threshold of 50\% diagonal length of $r_j$.
(iii) If $c'_i$ is an expansion of $c_i$ that physically supports $c_j$, then $c'_i$ must be not over-expanded.
To calculate $c'_i$, as shown in inset, we expand $c_i$ in the
direction parallel to floor and determine the extent of
\begin{wrapfigure}{r}{0.33\columnwidth}
	\vspace{-8pt}
	\begin{center}
\hspace{-30pt}
	\begin{overpic}[width=0.3\columnwidth]{\figures{expansion}}
	\put(10,87.0){$c_j$}
	\put(10,38.0){$c_j$}
	\put(65,69.0){$c_j$}
	\put(65,20.0){$c_j$}
	\put(83,69.0){$c_i$}
	\put(83,20.0){$c'_i$}
	\put(26,65.0){$c_i$}
	\put(26,16.0){$c'_i$}
    \put(10,-3.5){\small 3D cuboid}
    \put(60,-3.5){\small 2D projection}
    \end{overpic}
 	\end{center}
	\vspace{-8pt}
\end{wrapfigure}
expansion by finding a minimum bounding rectangle on the projective plane of floor that aligns to $r_i$ and encloses both $r_i$ and $r_j$.
Lastly, we say an object $v_i$ is {\em likely to occlude} $v_j$ if at least one of rays from viewpoint to vertices of frontal face of $c'_j$ intersects $c_i$, where $c'_j$ is the extrusion of $c_j$ to the floor.

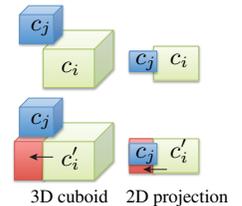
\mypara{Local refinement}
Most indoor objects retain consistent spatial relationships with respect to the room layout across various scenes (e.g., large furniture like bed and bookshelf are usually placed on the ground and against the walls).
We use the learned spatial constraints ($O_c$, $O_p$ and $O_s$) to guide the local refinement.
Given the latest object $v_i$ with label $l_i$ confirmed by user, we check the following conditions in turn and refine $c_i$ accordingly:
(i) If $l_i \in O_p$, we rotate $c_i$ such that its back face aligns to the nearest wall and re-estimate $c_i$ by fitting 3D points given the new orientation.
(ii) If $l_i \in O_c$, we extrude the back face of $c_i$ toward and touching the nearest wall, and it is not over-expanded.
(iii) If $l_i \in O_s$, we extrude the bottom face of $c_i$ to the floor.
(iv) If $v_i$ has a supporting parent $v_p$, we extrude/shrink the bottom face of $c_i$ to the top face of $c_p$.
Such local refinement, although simple, performs surprisingly well in improving the spatial relationships to the room and aligning the cuboid with underlying object image.
%
For example in Figure~\ref{fig:SG_refine}((a)-(b)) once the user confirmed the bed, system refined its orientation and dimension according to its learned spatial constraints to the room layout.

\mypara{Global refinement}
After the local refinement, we improve global inter-object relationships by examining the relationships of $v_i$ with respect to $v_j \in V_f$.
The local refinement of $c_i$, e.g., extruding and aligning to wall, might potentially improve its support relationships with $v_j \in V_f$.
Therefore, we reconstruct $G$ according to the new geometry of $c_i$ (if it is changed).
In addition, if $v_i$ is a ground object, i.e., $l_i \in O_s$, then it is likely to support or occlude nearby $v_j \in V_f$.
Note that object that is likely to be supported by $v_i$ could probably be occluded by $v_i$ as well, and we leave and resolve such ambiguity to later stage (e.g., pillow and dresser in Figure~\ref{fig:SG_refine}(b)).
Thus, we only search objects in $V_f$ that are likely to be occluded and not supported by $v_i$, and simply extrude these objects to the floor and reconstruct $G$.
%
%
Since it is not always 100\% correct to extrude potentially occluded objects, we allow user to 'undo' this operation if necessary.

Upon each local and global refinement, we update $G$ and use it as new input to label prediction followed by another user session.
The process iterates until user performs the `approve all' action at which moment all the labels are confirmed and our system performs a final refinement.
First, we traverse the support hierarchy of $G$ in a top-down, level-order and apply the local refinement to every visited object and reconstruct $G$ accordingly.
Second, for each object $v_i \in V_f$ we search $V_g$ for likely supporting parents $V_g'$, and find the object $v_j \in V_g'$ with highest $\mathcal{P}_s(l_j,l_i)$.
Given the strong priors by user, i.e., groundtruth labels, we resolve the ambiguity between support and occlusion as follows:
(i) If $l_i \in O_s$ and there is no or weak support evidence for $v_i$ ($V_g' = \{\emptyset\}$ or $\mathcal{P}_s(l_j,l_i) < 0.7$), we consider $v_i$ is a potential occluded ground object and thus extrude $v_i$ to the floor.
(ii) Otherwise, if $\mathcal{P}_s(l_j,l_i) \geq 0.7$, we physically expand $c_j$ to support $c_i$.
In Figure~\ref{fig:SG_refine}((c)-(d)), two floating pillows indicate a strong support relationships and thus affect the nearby bed which is expanded to support pillows.
Finally we reconstruct $G$ to obtain the final annotation for 3D structure and labels.

\mypara {Segment refinement}
While the initial cuboids are inferred from object segments, in reverse we exploit the refined cuboids to improve the quality of object segments.
Available depth simplifies this step.  We traverse $G$ in a bottom-up,
%
%
level-order and for each object $v_i \in V$ along with initial segments $S_i$, we search in $S$ for each image segment, $s_i$, that is enclosed by or intersects with the convex hull of the image-space projection of $c_i$.
Then, if 80\% of 3D points of $s_i$ are inside the volume of $c_i$ we absorb $s_i$ into $S_i$ and label it with $l_i$.
Otherwise, if $s_i$ was originally in $S_i$, we unlabeled and remove it from $S_i$.
%
%

\begin{figure*}[!t]
    \centering
    \includegraphics[width=\textwidth]{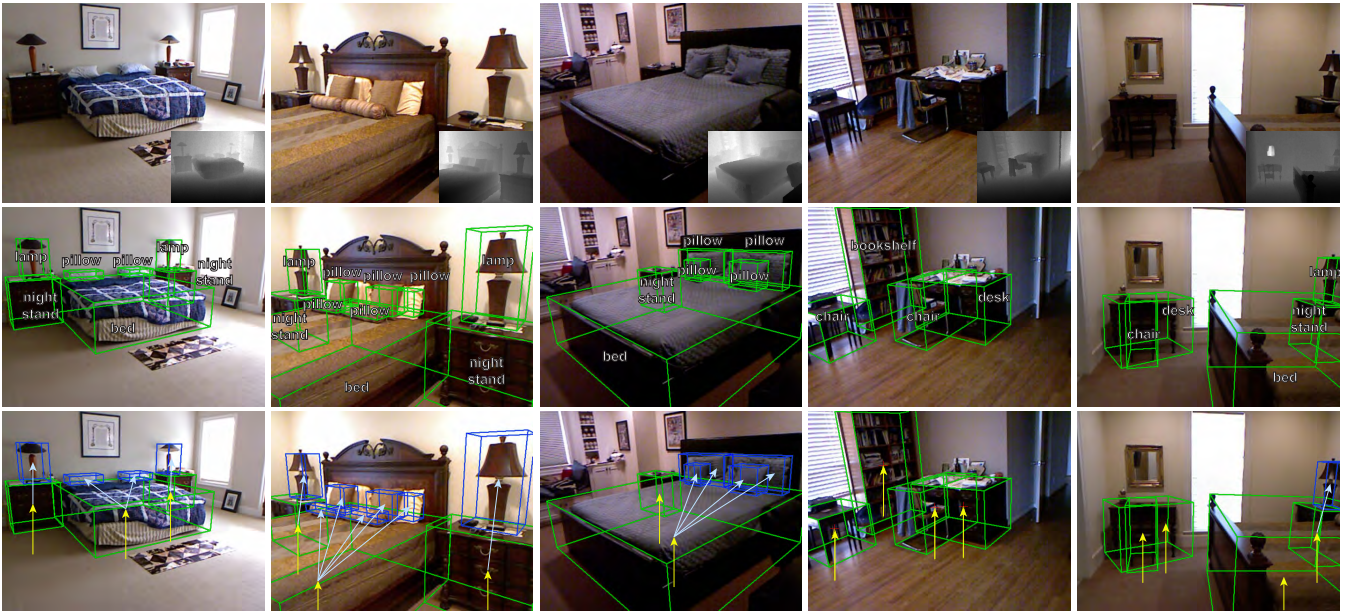}
    \caption{Five sample annotated scenes from our experimental studying. The top row shows input {\rgbd} images, the middle row presents the annotated data with object labels and 3D structure of the scene, and the inferred supporting hierarchy is shown in bottom row. The yellow and light blue arrows indicate support relationships with floor and other objects, respectively.}
    \label{fig:results}
    \vnudge
\end{figure*}

\section{Evaluation}
\label{sec:eval}
We tested our system on the benchmark NYU-Depth dataset~\cite{silberman:12:indoor} which consists of 1449 {\rgbd} images that covers 26 scene classes and 894 object categories, where each image contains detailed object segments and labels annotated by users.
We conducted an experiment to quantantively and qualitatively evaluate the performance of SmartAnnotator, demonstrating that it effectively learns richer priors as user processes more scenes and progressively simplifies the annotation process.

\mypara{Experimental setting}
We randomly picked 10 images from the `bedroom' scene class to bootstrap the learning process and used 9 commonly seen object categories in the bedroom scene
as our target labels.
Specifically, we choose `bed', `sofa', `dresser', `night stand', `desk', `chair', `bookshelf', `lamp', and `pillow.' 
While in the label prediction session, the system displays 6 suggestions for each object.
The objective of the experiment was to annotate another 126 bedroom scenes, chosen at random, from the NYU-Depth dataset, using the learned priors.
As preprocessing, we parsed and constructed SGs for the target images based on the annotated object segments.
To avoid distractions during the annotation process, objects that are not belong to target categories are filtered out in advance.
We designed an incremental learning scheme by uniformly dividing the experiment into 7 trials with each trial contains 18 scenes.
We arranged the images such that the complexity of scenes, in terms of number of objects and occurrence of support relationships, are similar across trials to prevent measurement bias.
%
Before starting the experiment, users were given a tutorial ($\sim$5 mins) to be familiar with the interface and the flow of annotation process.
%
Figure~\ref{fig:results} shows 5 annotated scenes from the experiment.
Please refer to supplementary material for a complete set of tested scenes.

\mypara{Performance evaluation}
We first evaluated the performance of our system in terms of
manual efforts and timing in annotating phase.
%
The major labors are from manual actions performed in user session to refine suggestions from system (i.e., `Confirm', `Re-order' and `Type' actions in Section~\ref{sec:user}).
Therefore, we quantized the measurement by counting the frequency of user performing each action among all objects in a trial.
We further evaluated the performance of label prediction and suggestion algorithm by computing a `Top-3-Hit' ratio that indicates the groundtruth label is among the top 3 suggestions.
%
For timing, we compared the averaged elapsed time of annotating process with and without the assistance of our system.
For a na\"{\i}ve low-level annotating, we prepared another trial and measured the timing of user one-by-one typing in the label for each object.

As shown in Figure~\ref{fig:quanty_eval}(a), although our system bootstraps from a weak model, overall performance improves as user processes more scenes in the subsequent trials.
This shows the effectiveness of our system as it incrementally gets richer priors which in turn benefits label prediction and suggestion as shown in the `Top-3-Hit' curve.
As a result, the annotation process is progressively simplified requiring lesser average annotation time, which, not surprisingly, is much better than the na\"{\i}ve manual labeling as shown in Figure~\ref{fig:quanty_eval}(b).
%
Also note that performance degrades slightly in the last two trials as can be expected from an incremental learning scheme where errors accumulate overtrials and can damage later priors learning.
%

\mypara{Quality of support relationships}
We compared the quality of our inferred support relationships, both before and after the structure graph refinement, with groundtruth data.
To generate groundtruth SGs for 126 scenes, we first constructed initial SG for each image and then manually refined its edge structure.
Given a target SG $G$ and its groundtruth SG $G_{gt}$, we measure error between $G$ and $G_{gt}$ by counting the number of required edge insertion and deletion to transform from graph $G$ to $G_{gt}$.
Note that since every object is at least supported by one another object, including the floor, we counted support error rate in a trial with respect to total number of objects among 126 groundtruth SGs.
Figure~\ref{fig:quanty_eval}(c) shows the error rate of each trial both from the initial and refined SG, indicating that our structure graph refinement could effectively improve the quality of support relationships.

\begin{figure}[b!]
\vnudge
    \centering
    \includegraphics[width=\columnwidth]{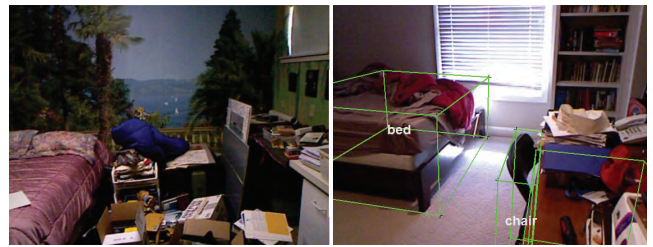}
    \caption{Limitations. (Left) Scene with missing floor. (Right) Complex interaction between two non-convex objects (chair and desk).}
    \label{fig:limit}
\end{figure}

\textbf{Limitations }
of our system includes:
(i) It is unable to handle images where the floor is completely invisible and hence fails to infer proper floor segments (see Figure~\ref{fig:limit}(left)).
In such case, all the ground objects are floating and partially occluded, and the local refinement to fix ground cuboids is disabled due to the lack of reference floor.
(ii) The simplified 3D abstraction using cuboids could not capture complex interaction between non-convex objects (e.g., a chair is tucked inside a desk), and therefore introducing erroneous measurement in cuboid dimension (see Figure~\ref{fig:limit}(right)).
A detailed 3D representation is required to obtain more accurate  information.
%
(ii) Without employing an introspection, our model can suffer from learning erroneous 3D structures from earlier trials and thus damage the later ones.

\section{Conclusion}
We presented an interactive tool for smart annotation of {\rgbd} indoor scene images with both image and scene level contents including image segments, segment labels, and structure of the scene.
In the learning phase, we design a novel scene labeling to train geometric and structural priors by reasoning on 3D structure of the scene inferred from the depth data.
In the annotating phase, system assists user in performing laboring tasks of computing segmentations, predicting the object labels using the learned priors, and inferring the 3D structure of scene, while user play.
User simply supervises the process by progressively accepting suggestionf from the system which in turn updates the labels and structure in response to user.
Such pattern iterates until user approves all labels.
In the near future, we plan to both release source codes of our tool and deploy the system as a web-based application.

Several interesting challenges lies ahead:
(i)~Incorporating an introspection scheme to alleviate the error accumulation issue which involve re-annotating scenes in earlier trials using the latest models would be an interesting future direction.
(ii)~While we only tested the system on a particular set of images and objects (e.g., bedroom and 9 object labels), our system is flexible to be extended to support multi-scene classes and arbitrary object categories.
We plan an extension of learning separated models for different scene classes while in annotation process, we exploit a scene classifier~\cite{Quattoni:09:indoor} to choose target model and dynamically extend the categories when user typing a novel label.
However, na\"{\i}vely extending categories might potentially pollute the learning process and thus requires more thoughts.
(iii)~In the learning phase, we have focused on geometry/texture and support relationships.
Other potential contextual relationships, such as relative position and orientation between two objects (e.g., nightstand is often beside the bed) could potentially be used. 
(iv)~Finally, to account for more complicated structural relationships,
one needs a detailed 3D model of object, especially for objects of non-convex shape, in order to estimate accurate contact information.
%
%
However, how to automate the process remains to be investigated. 

\label{sec:conclusion} 


\bibliographystyle{acmsiggraph}
\bibliography{RGBD_Annotate}

\end{document}